\documentclass[]{elsarticle}
\usepackage[utf8]{inputenc}
\usepackage[colorinlistoftodos]{todonotes}
\usepackage{multirow}
\usepackage{algorithm} 
\usepackage{stackrel} 
\usepackage{mathtools} 
\usepackage{booktabs} 
\usepackage{tabularx} 
\usepackage[normalem]{ulem}
\usepackage{xcolor}
\usepackage{subcaption}
\usepackage{graphicx}

\begin{document}

\title{
CI-dataset and DetDSCI methodology for detecting too small and too large critical infrastructures in satellite images: Airports and electrical substations as case study
}

\author[1]{Francisco Pérez-Hernández\corref{cor1}} \ead{fperezhernandez@ugr.es}
\address[1]{Andalusian Research Institute in Data Science and Computational Intelligence, University of Granada, 18071 Granada, Spain}
\author[1]{Jose Rodríguez-Ortega} \ead{jrodriguez98@correo.ugr.es}
\author[1]{Yassir Benhammou} \ead{benhammouyassir2@gmail.com}
\author[1]{Francisco Herrera} \ead{herrera@decsai.ugr.es}
\author[1]{Siham Tabik} \ead{siham@ugr.es}
\cortext[cor1]{Corresponding author}

\begin{abstract}
	The detection of critical infrastructures in large territories represented by aerial and satellite images is of high importance in several fields such as in security, anomaly detection, land use planning and land use change detection. However, the detection of such infrastructures is complex as they have highly variable shapes and sizes, i.e., some infrastructures, such as electrical substations, are too small while others, such as airports, are too large. Besides, airports can have a surface area either small or too large with completely different shapes, which makes its correct detection challenging. As far as we know, these limitations have not been tackled yet in previous works. This paper presents (1) a smart Critical Infrastructure dataset, named CI-dataset, organised into two scales, small and large scales critical infrastructures and (2) a two-level resolution-independent critical infrastructure detection (DetDSCI) methodology that first determines the spatial resolution of the input image using a classification model, then analyses the image using the appropriate detector for that spatial resolution. The present study targets two representative classes, airports and electrical substations. Our experiments show that DetDSCI methodology achieves up to 37,53\% F1 improvement with respect to Faster R-CNN, one of the most influential detection models.
\end{abstract}


\maketitle

\section{Introduction}

Critical infrastructures are a type of human land use that are essential for the functioning of a society and economy \cite{oshri2018infrastructure, xiao2017airport, yang2010bag}. Any threat to these facilities can cause severe problems. Examples of critical infrastructures include airports, electrical substations and harbours among others. The detection of this type of infrastructures in high resolution ortho-images is of paramount importance in several fields such as security, land use planning and change detection \cite{carranza2019framework, guidici2017one, liu2019integration, zhang2019joint}.

Currently, deep CNNs have been largely used in the classification of high resolution ortho-images \cite{cheng2017remote, christie2018functional, yang2010bag} as they achieve good accuracies specially in distinguishing infrastructures of similar scales in images of the same size and same spatial resolution. Nevertheless, the detection of critical infrastructures with dissimilar sizes and scales, e.g., electrical substations, which usually cover a surface area of the order of hundreds m$^2$, versus airports, which can cover from few to hundreds km$^2$, is still challenging.

Such task is addressed using remote sensing data and deep Convolutional Neural Networks (CNNs). Remote sensing data are high resolution ortho-images that can be obtained from Unmanned Aerial Vehicle (UAV) (captured at height $<30km$ and covers from 0,1 to 100$Km^2$), planes (at height $<30km$ and covers from 10 to 100$Km^2$) or satellites ($>150km$ 10 to 1000 Km$^2$) \cite{xiang2018mini}. Obtaining large amounts of this type of data is expensive. Fortunately, several sources, such as Google Earth\footnote{Google Earth: https://earth.google.com/web} and Bing Maps\footnote{Bing Maps: https://www.bing.com/maps}, allow downloading aerial and satellite images freely for the academic community. In spite of this, most existing land use datasets are prepared for training classification models and do not include annotations for training detection models. 

This paper presents two-level deep learning Detection for Different Scale Critical Infrastructures (DetDSCI) methodology in ortho-images. We reformulate the problem of detecting critical infrastructures in ortho-images into two sub-problems, the detection of too small and too large scale critical infrastructures. DetDSCI methodology detects the type of infrastructure independently of its scale and consists of two stages:

\begin{itemize}
	\item The first stage is based on a spatial resolution classification model that analyses the $2000\times2000$ pixels input image to estimate its zoom level and hence determine the detector to be used in the next stage. 
	\item The second stage includes two expert detectors, one for small and the other for large critical infrastructures. Once the zoom level of the input image is determined by the first stage, the selected detector will analyse that input image according to its spatial resolution.
\end{itemize}

Addressing the detection of too small and too large scale critical infrastructures in remote sensing images independently on the spatial resolution can offer better performance. Our study targets two representative critical infrastructures, namely airports and electrical substations. As there are no public detection datasets that include both categories of critical infrastructures, we carefully built a specialised dataset, Critical Infrastructures dataset (CI-dataset). CI-dataset is organised into two subsets, Small Scale Critical Infrastructure (CI-SS) dataset with electrical substation class and Large Scale Critical Infrastructure (CI-LS) dataset with airport class. 

The main contributions of this paper can be summarised as follows:

\begin{itemize}
	\item Differently to the traditional process adopted for building most datasets, we followed a dynamic process for constructing the high quality CI-dataset organised into two scales, CI-SS for small scale critical infrastructures and CI-LS for large scale critical infrastructures. This process can be used to include more types of infrastructures. CI-dataset is available through this link\footnote{CI-dataset: https://dasci.es/transferencia/open-data/ci-dataset/}.
	\item We present DetDSCI methodology, a two-stages deep learning detection for dissimilar scale critical infrastructures in ortho-images. DetDSCI methodology first determines the spatial resolution of the input image then analyses it according to its spatial resolution using the appropriate expert detector. This methodology overcomes the baseline detectors trained on our high quality dataset.
\end{itemize}

This paper is organised as follows. First, a comprehensive review of related works is provided in Section \ref{sec_related_works}. Our DetDSCI methodology is presented in Section \ref{sec_detdsci}. The dynamic process of building our CI-dataset is provided in Section \ref{sec_ci_dataset}. The experimental analysis carried out for the construction of CI-dataset and the evaluation of DetDSCI methodology are given in Section \ref{sec_experimental_study}. Finally, conclusions and future works are given in Section \ref{sec_conclusions}.

\section{Related works} \label{sec_related_works}

Related works that apply deep learning on remote sensing data can be broadly divided into two types, top-down and bottom-up works: 

\begin{itemize}
	\item Top-down works, first build a large dataset with an important number of object-classes, mainly objects that can be recognised from remote sensing images, e.g., vehicles or soccer stadiums. Then, analyse these images using a deep learning classification or detection models \cite{cheng2017remote, cheng2014multi, cheng2016learning, christie2018functional, lam2018xview, li2020object, xia2018dota, yang2010bag}. 
	\item Bottom-up works focus on solving one specific problem that involves one or few object classes, e.g., airports \cite{budak2018deep, cai2017airport, li2019remote, xu2018end, zhang2017airport}, trees \cite{bjerreskov2021classification, flood2019using, guirado2020tree, safonova2021olive} and whales \cite{guirado2019whale}. 
\end{itemize}

Our work belongs to the second category as our final objective is to build a good detector of two specific critical infrastructures, namely, airports and electrical substations. This section provides a brief summary of the current general datasets that include some critical infrastructures, the so-called top-down works (Section \ref{rw_datasets}) then reviews the deep learning approaches used in bottom-up works (Section \ref{rw_works}).

\subsection{Top-down works} \label{rw_datasets}

Most databases provided by top-down works are multi-class datasets that include some critical infrastructures, annotated for the task of image classification, which limits their usefulness. See summary in Table \ref{table_datasets_rw} where only a few datasets are prepared for the task of detection.

\begin{table}[H]
	\caption{Characteristics of general datasets that include some critical infrastructures.}
	\label{table_datasets_rw}
	\centering
	\resizebox{\textwidth}{!}{
	\begin{tabular}{lrrrrrr}
	\toprule
	Dataset & \begin{tabular}[c]{@{}r@{}}\#Classes\\ (\#Infrastructure)\end{tabular} & \begin{tabular}[c]{@{}r@{}}\#Images\\ (\#Instances)\end{tabular} & \begin{tabular}[c]{@{}r@{}}\#Image\\ width\end{tabular} & Source & Resolution & Annotation \\ \midrule
	LULC\cite{yang2010bag} & 21 (7) & 2100 (2100) & 256 & National Map & 30cm & Classification \\
	\begin{tabular}[c]{@{}l@{}}NWPU\\ RESISC45\cite{cheng2017remote}\end{tabular} & 45 (13) & 31500 (31500) & 256 & Google Earth & 20cm-30cm & Classification \\
	fMoW\cite{christie2018functional} & 62 (25) & 523846 (132716) & N/A & OpenStreetMap & 31cm-1.6m & Classification \\ \midrule
	\begin{tabular}[c]{@{}l@{}}NWPU\\ VHR-10\cite{cheng2014multi}\end{tabular} & 10 (4) & 800 (3651) & $\sim$1000 & Google Earth & 15cm-12m & Horizontal BB \\
	xView\cite{lam2018xview} & 60 (9) & 1400 (1000000) & 3000 & DigitalGlobe & 31cm & Horizontal BB \\
	DIOR\cite{li2020object} & 20 (11) & 23463 (192472) & 800 & Google Earth & 30cm-50cm & Horizontal BB \\
	DOTA\cite{xia2018dota} & 15 (6) & 2806 (188282) & 800$\sim$4000 & Google Earth & 15cm-12m & Oriented BB \\
	\bottomrule
	\end{tabular}
	}
\end{table}

For example, in \cite{yang2010bag}, the authors created LULC dataset organised into 21 classes. Each class contains $100$ images of size $256\times256$ pixels. The authors in \cite{cheng2017remote} provide a dataset named NWPU-RESISC45. This dataset is composed of $31.500$ images of $256\times256$ pixels, in 45 classes with $700$ images in each class. NWPU-RESISC45 includes images with a large variation in translation, spatial resolution, viewpoint, object pose, illumination, background, and occlusion. Besides, it has high within-class diversity and between-class similarity. Functional Map of the World (fMoW) \cite{christie2018functional} is a dataset containing a total of $523.846$ images with a spatial resolution of $0,31$ and $1,60$ meters per pixel. It includes $62$ classes with $132.716$ instances from OpenStreetMap. These datasets are prepared for the image classification task and hence they are not useful for the detection task.

Examples of datasets prepared for the task of object detection are NWPU VHR-10, xView, DIOR and DOTA. NWPU VHR-10 dataset \cite{cheng2014multi} is organised into $10$ classes, each class contains 800 images of width 1000 pixels. It contains mainly small scale objects such as airplane, ship, storage tank, baseball diamond, tennis court, basketball court, ground track field, harbour, bridge, and vehicle. Authors on \cite{lam2018xview} presented xView dataset for detecting 60 object-classes with over 1 million instances. These classes are focused on vehicles and small scale objects and the images have a width of 3000 pixels. DIOR, a new dataset was published on \cite{li2020object}, where 23463 images and 192472 instances covered 20 object classes. DIOR dataset has a large range of object size variations and is focused on detection with a width on the images of 800 pixels. DOTA dataset \cite{xia2018dota} is composed of 15 classes of small scale objects with $2.806$ images from Google Earth where the total instances are $188.282$. The size of the images is between $800$ and $4.000$ pixels, and they are labelled with oriented bounding boxes. Although the last four datasets are prepared for the task of object detection, they do not focus on any specific problem as they are all types of visible objects from space. In addition, none of these datasets includes electrical substations and only DIOR includes the airport category.

\subsection{Bottom-up works} \label{rw_works}

A large number of bottom-up works focus on improving the detection of airports. In \cite{zhang2017airport}, the authors propose a method using CNNs for airport detection on optical satellite images. The proposed method consists mainly of three steps, namely, region proposal, CNN identification, and localisation optimisation. The model was tested on an image data set, including 170 different airports and 30 non-airports. All the tested optical satellite images were collected from Google Earth with a resolution of $8m\times8m$ and a size of about $3000\times3000$ pixels. The method proposed in \cite{budak2018deep} first detects various regions on RSIs, then uses these candidate regions to train a CNN architecture. The sizes of the airport images were $3000\times2000$ pixels with a resolution of 1m. A total of 92 images were collected. In \cite{cai2017airport}, the authors developed a hard example mining and weight-balanced strategy to construct a novel end-to-end convolutional neural network for airport detection. They designed a hard example mining layer to automatically select hard examples by their losses and implement a new weight-balanced loss function to optimise CNN. The authors in \cite{xu2018end} proposed an end-to-end airport detection method based on convolutional neural networks. Additionally, a cross-optimisation strategy has been employed to achieve convolution layer sharing between the cascade region proposal networks and the subsequent multi-threshold detection networks, and this approach significantly decreased the detection time. Once the airport is detected, they use an airplane detector to obtain these instances. To address the insufficiency of traditional models in detecting airports under complicated backgrounds from remote sensing images, authors in \cite{li2019remote} proposed an end-to-end remote sensing airport hierarchical expression and detection model based on deep transferable convolutional neural networks.

\section{DetDSCI methodology: Two-level deep learning Detection for Different Scale Critical Infrastructure methodology in ortho-images} \label{sec_detdsci}

\begin{figure}[H]
	\centering
	\includegraphics[width=0.95\textwidth]{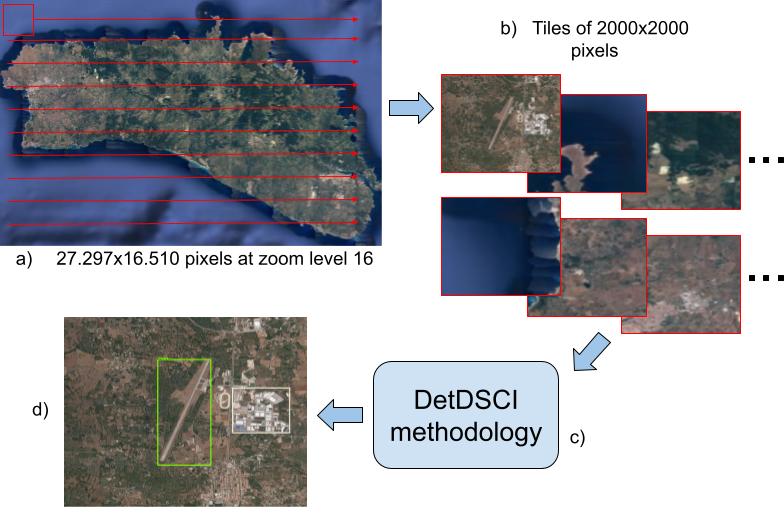}
	\caption{DetDSCI Methodology detection applied to the island of Menorca (Spain). (a) A sliding window processing approach. (b) Obtained $2000\times 2000$ pixels crops. (c) DetDSCI methodology applied to each crop. (d) Output image with detection results.} 
	\label{fig_sliding}
\end{figure}

This section presents DetDSCI methodology which aims at addressing the detection of airports and electrical substations of very dissimilar sizes and shapes in large areas represented by satellite images, see illustration in Figure \ref{fig_sliding}. We define two broad ranges of spatial resolutions also called zoom levels, see correspondence between zoom level and spatial resolution in Table \ref{tab_zoom_level}. The first range includes zoom levels in [14,17] and the second range includes zoom levels in [18,23]. These intervals have been selected experimentally as described in the next section.

\begin{table}[H] 
	\centering
	\caption{The correspondence between spatial resolution and zoom level.}
	\label{tab_zoom_level}
	\footnotesize
	\begin{tabular}{cc|cc}
	\toprule
	\multicolumn{2}{c|}{Large critical infrastructures} & 	\multicolumn{2}{c}{Small critical infrastructures}\\
	Zoom level & Spatial resolution($m^2$/pixel) & Zoom level & Spatial resolution($m^2$/pixel) \\ 
	\midrule
	14 & 6.2 & 18 & 0.39 \\
	15 & 3.1 & 19 & 0.19 \\
	16 & 1.55 & 20 & 0.10 \\ 
	17 & 0.78 & 21 & 0.05 \\
	& & 22 & 0.02 \\
	& & 23 & 0.01 \\ \bottomrule
	\end{tabular}
\end{table} 

\begin{figure}[H]
	\centering
	\includegraphics[width=0.6\textwidth]{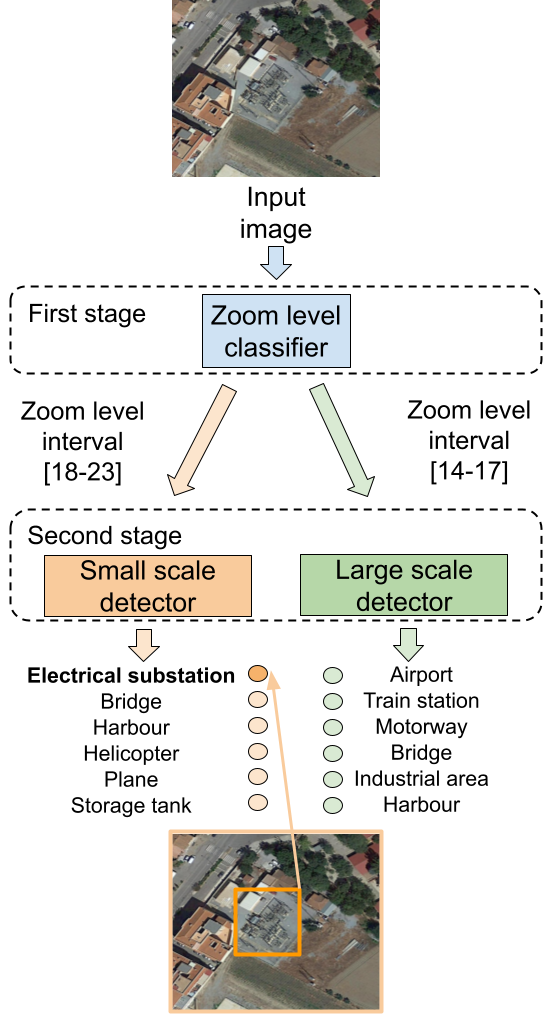}
	\caption{DetDSCI methodology.} \label{figure_DetDSCI}
\end{figure}

To reduce the number of false positives due to the differences in different zoom levels, DetDSCI methodology first distinguishes between the two zoom level ranges then applies the corresponding detector according to the spatial resolution of each input image. In particular, DetDSCI is actually a two stages pipeline as illustrated in Figure \ref{figure_DetDSCI}. The first stage determines whether the input image belongs to the first or second zoom levels interval. Depending on the selected zoom level interval, the second stage analyses that image using the specialised detector on that specific group of critical infrastructures.

\subsection{Stage 1: Estimating the spatial resolution of the input image}

\begin{figure}[H]
	\centering
	\includegraphics[width=0.95\textwidth]{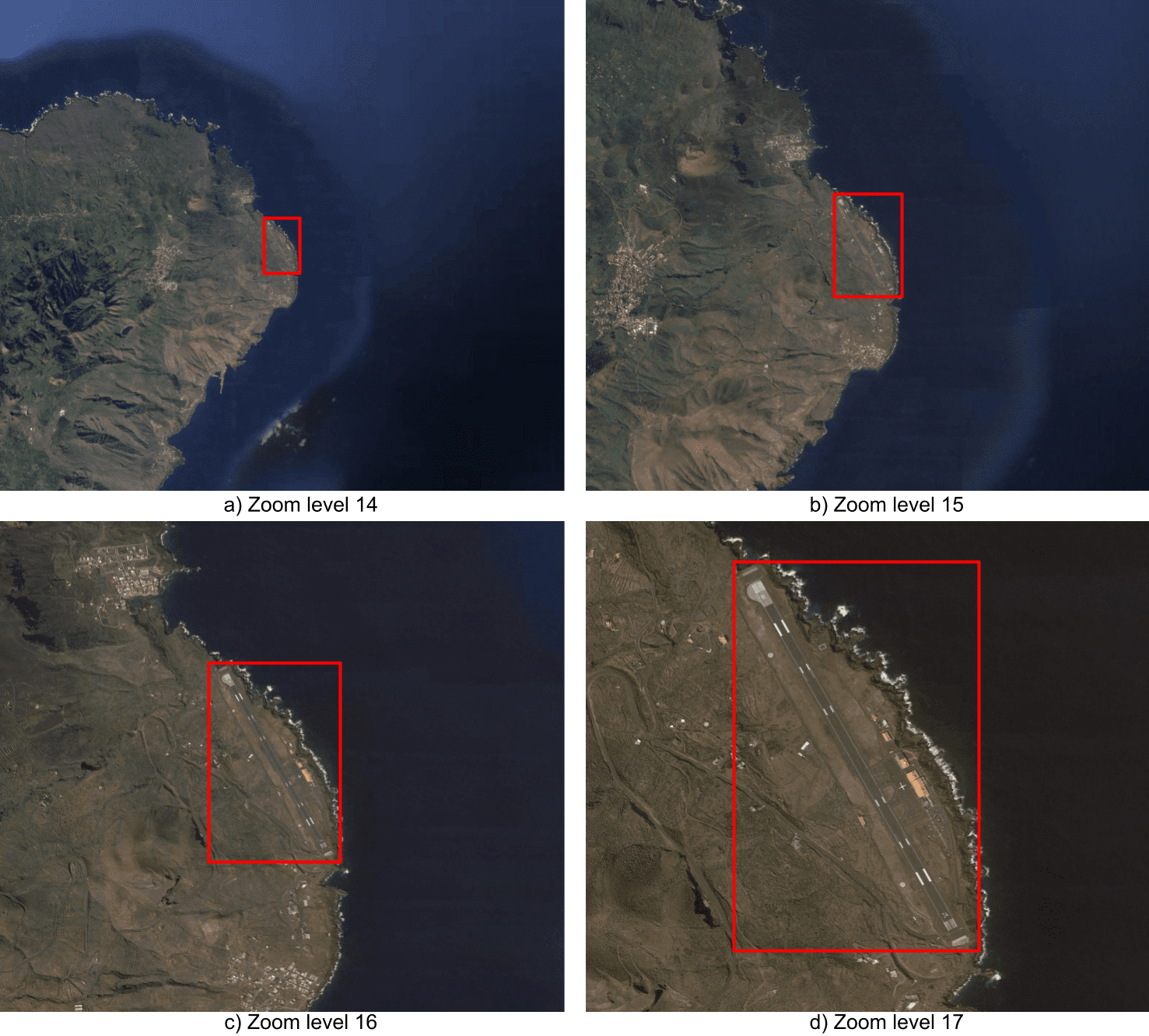}
	\caption{Four images of El Hierro airport (latitude: 27.81402${^o}$N, longitude: -17.88518${^o}$W, Canary Islands, Spain) with zoom levels 14(a), 15(b), 16(c) and 17(d), obtained from Google Maps.} 
	\label{fig_zoom_largescale}
\end{figure}

\begin{figure}[H]
	\centering
	\includegraphics[width=0.95\textwidth]{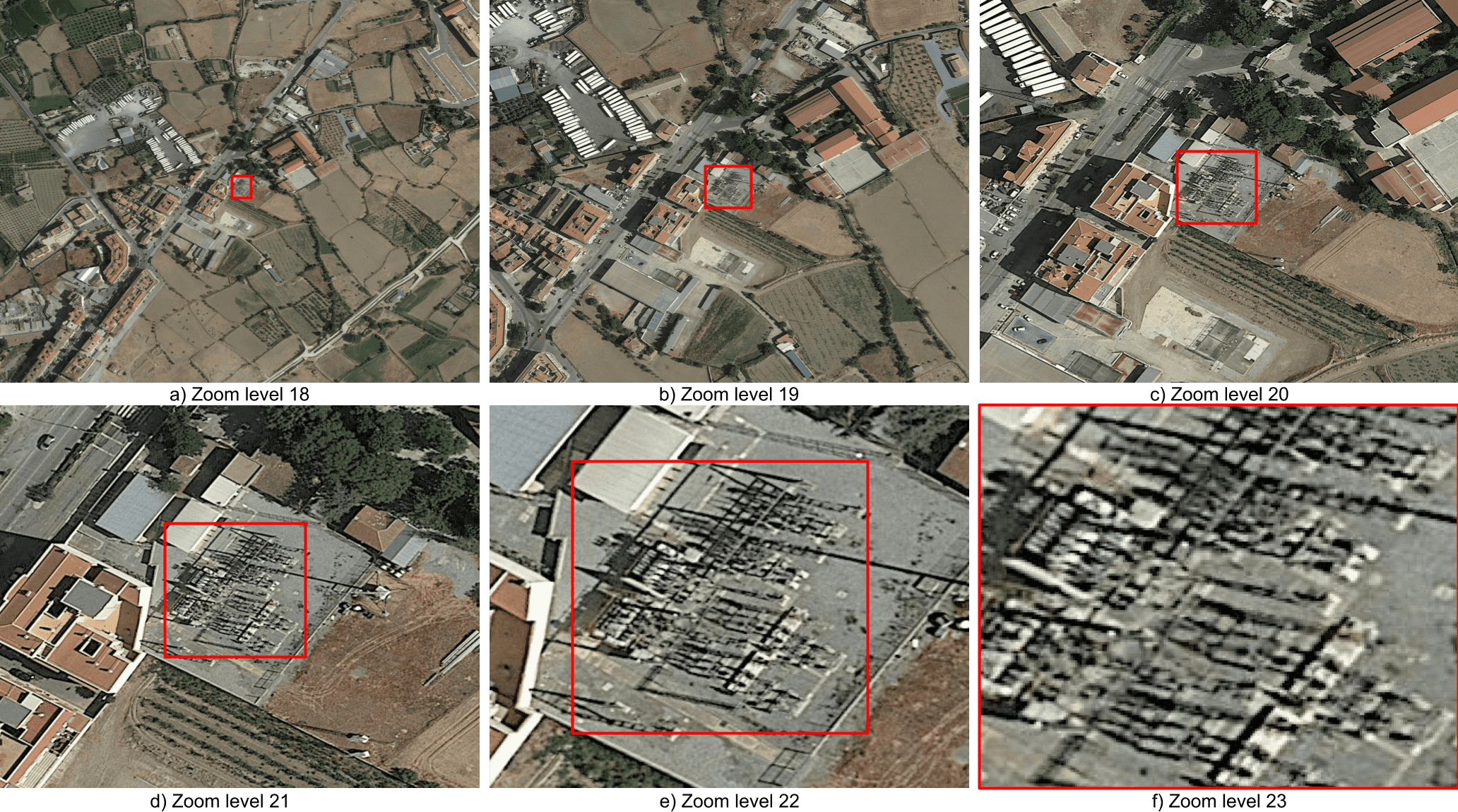}
	\caption{Six images of Guadix electrical substation (latitude: 37.30853${^o}$N, longitude: -3.12997${^o}$W, Granada, Spain) with zoom levels 18(a), 19(b), 20(c), 21(d), 22(e) and 23(f), obtained from Google Maps.} 
	\label{fig_zoom_smallscale}
\end{figure}

To distinguish between too large and too small critical infrastructures, we consider two zoom levels intervals, [14,17] and [18,23]. Too large infrastructures can be visually recognised in $2000\times 2000$ pixels images of zoom levels 14, 15, 16 and 17. See an example in Figure \ref{fig_zoom_largescale}. While, too small scale infrastructures can be visually recognised in $2000\times 2000$ pixels images of zoom levels 18, 19, 20, 21, 22 and 23. See an example in Figure \ref{fig_zoom_smallscale}.

The first stage of DetDSCI distinguishes between these two intervals, large [14,17] and small [18,23] zoom levels interval. This stage is based on a binary classification model that analyses the input image to determine its zoom level interval and hence determines the most appropriate detector to be used in the second stage.

\subsection{Stage 2: Detection of critical infrastructures}

\begin{figure}[H]
	\centering
	\includegraphics[width=\textwidth]{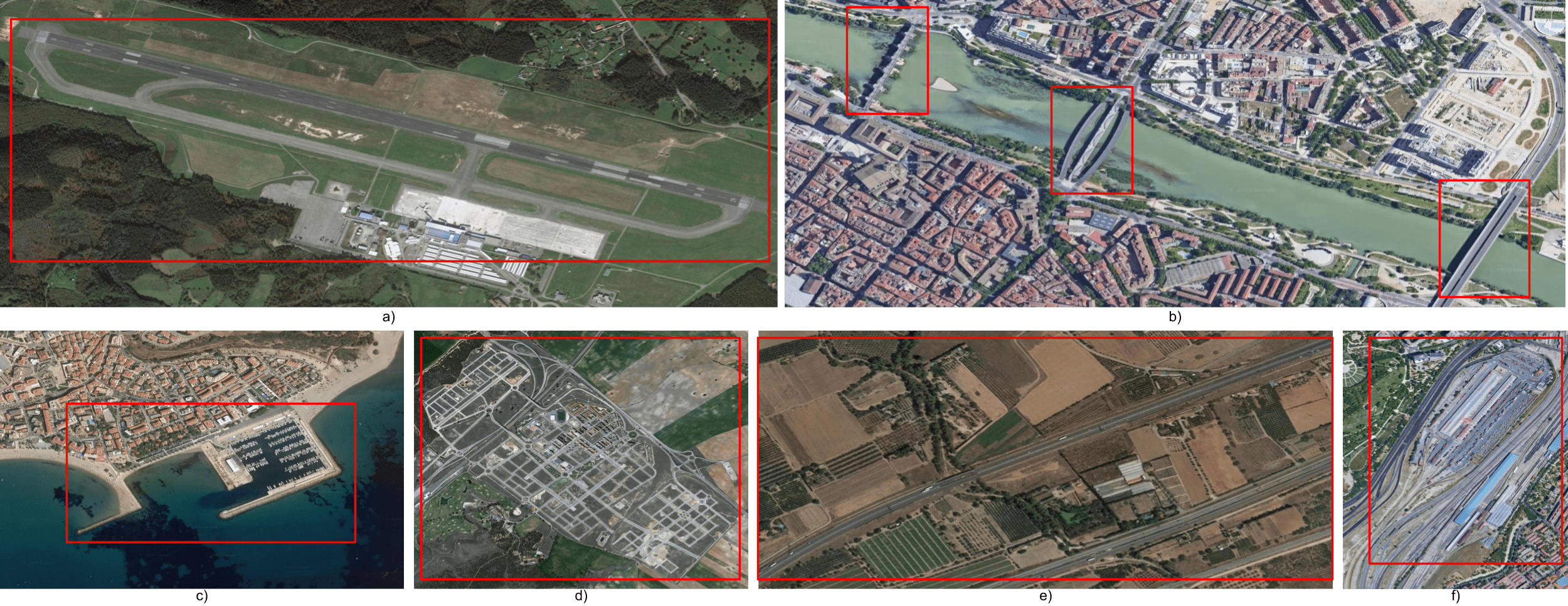}
	\caption{Examples of the classes considered by the large infrastructure detection model, left to right: airport(a), bridges(b), harbour(c), industrial area(d), motorway(e) and train station(f).} \label{fig_classes_largescale}
\end{figure}

\begin{figure}[H]
	\centering
	\includegraphics[width=0.95\textwidth]{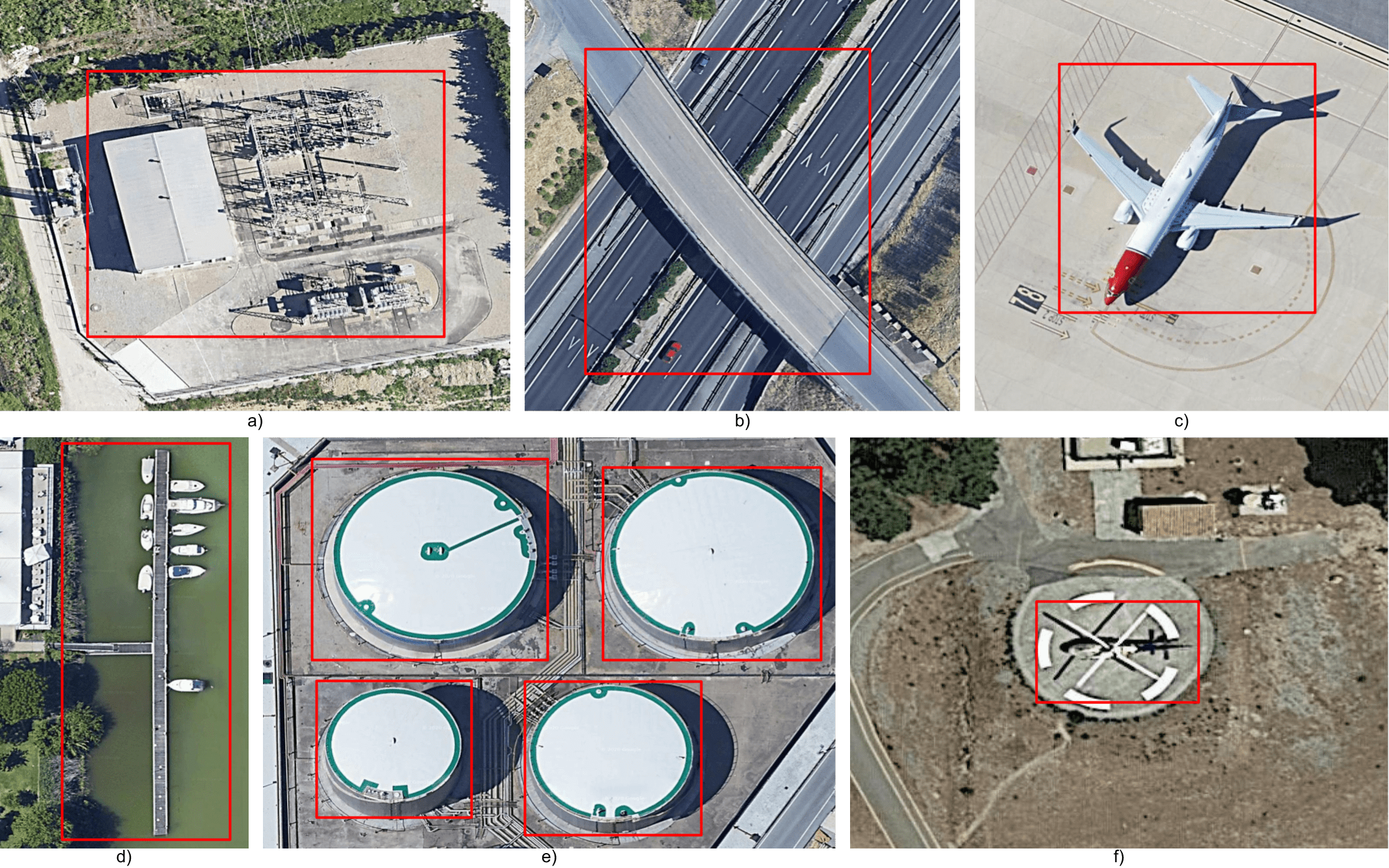}
	\caption{Examples of the classes considered in the small infrastructure detection model, left to right: electrical substation(a), bridge(b), plane(c), harbour(d), storage tanks(e) and helicopter(f).} \label{fig_classes_smallscale}
\end{figure}

The zoom level interval estimated in the first stage will be used to guide the selection of the detector in the second stage. In particular, this stage is based on two detection models:

\begin{itemize}
	\item The first detection model is applied to large scale infrastructures. It considers six infrastructure classes, namely airport, bridge, harbour, industrial area, motorway and train station. Figure \ref{fig_classes_largescale} shows examples of these classes.
	\item The second detection model is applied to small scale infrastructures. It considers six classes, namely electrical substation, bridge, plane, harbour, storage tank and helicopter. Figure \ref{fig_classes_smallscale} shows examples of these classes. 
\end{itemize}

It is worth mentioning that the inclusion of new classes in both detectors was based on the preliminary experimental study explained in the next section.

\section{CI-dataset construction guided by the performance of Faster R-CNN} \label{sec_ci_dataset}

It is well known that building good quality models requires good quality datasets, also called smart data \cite{luengo2020big}. The concept of smart data includes all pre-processing methods that improve value and veracity of data. In the context of object detection, usually training datasets are first built then analysed using machine learning models. This classical procedure is suitable only when the involved objects are of similar sizes and can be correctly identified at the same spatial resolution.

To overcome these limitations, we built the critical infrastructures dataset, CI-dataset, guided by the performance of one of the most robust detectors, namely Faster R-CNN. We organised CI-dataset into two subsets, one for small scale, CI-SS and the other one for large scale, CI-LS critical infrastructures. The construction process of both subsets is dynamic and guided by the performance of Faster R-CNN detection model on the electrical substation class for CI-SS and the airport class for CI-LS. This section describes the construction process used to obtain the final high-quality CI-dataset for detecting electrical substations and airports.

The dynamic process guided by the detection model is based on three main steps:

\begin{itemize}
	\item \textbf{Step 1: Constructing the initial set for each target class:} First, we selected the combination of zoom levels at which the airports and the electrical substations can be recognised by the human eye. Then, we downloaded images for each one of these two classes with different zoom levels. Afterwards, we selected the most suitable zoom levels combination guided by the performance of Faster R-CNN.
	\item \textbf{Step 2: Extending the dataset with more object classes:} We analysed all the object classes that can be confused with the target class and hence can cause false positives (FP). All these potential FP are obtained from public datasets and included in our CI-dataset. Then the performance of the model is analysed to select the final object classes to be included. 
	\item \textbf{Step 3: Further increasing the size of the training set:} We increased the number of instances of the final classes in the training set using new images from Google Maps. 
\end{itemize}

For simplicity, we named the three different versions of the training, test datasets and detection model according to the construction step as described in Table \ref{table_sumary_db}. At the end of this process, we obtained the final CI training and test datasets.

\begin{table}[H]
	\centering
	\footnotesize
	\caption{The names of the training and test subsets of the CI-dataset and the corresponding detection model created at each step of the process.}
	\label{table_sumary_db}
	\resizebox{\textwidth}{!}{
	\begin{tabular}{llll} \toprule
	& Train & Test & Detection model \\ \midrule
	Step 1 & CI-SS\_train\_alpha & CI-SS\_test\_alpha & CI-SS\_Det\_alpha \\
	Step 2 & CI-SS\_train\_beta & CI-SS\_test\_stable & CI-SS\_Det\_beta \\
	Step 3 & CI-SS\_train\_stable & CI-SS\_test\_stable & CI-SS\_Det\_stable \\ \midrule
	Step 1 & CI-LS\_train\_alpha & CI-LS\_test\_alpha & CI-LS\_Det\_alpha \\
	Step 2 & CI-LS\_train\_beta & CI-LS\_test\_stable & CI-LS\_Det\_beta \\
	Step 3 & CI-LS\_train\_stable & CI-LS\_test\_stable & CI-LS\_Det\_stable \\ \bottomrule
	\end{tabular}}
\end{table}

\subsection{Step 1: Constructing the initial set for each target class}

The first process is to carefully select the zoom levels at which the considered objects fit in a $2000\times2000$ pixels image and can be recognised by the human eye. Ortho-images of this size can capture small scale critical infrastructures within 18 to 23 zoom levels (see Figure \ref{fig_classes_smallscale}) and large scale critical infrastructures within 14 to 17 zoom levels (see Figure \ref{fig_classes_largescale}). For building CI-dataset, we used two services to visualise then download images from Google Maps, namely, SAS Planet\footnote{SAS Planet: //www.sasgis.org/} and Google Maps API\footnote{Google Maps API: //https://cloud.google.com/maps-platform}. 

\begin{figure}[H]
	\centering
	\includegraphics[width=0.95\textwidth]{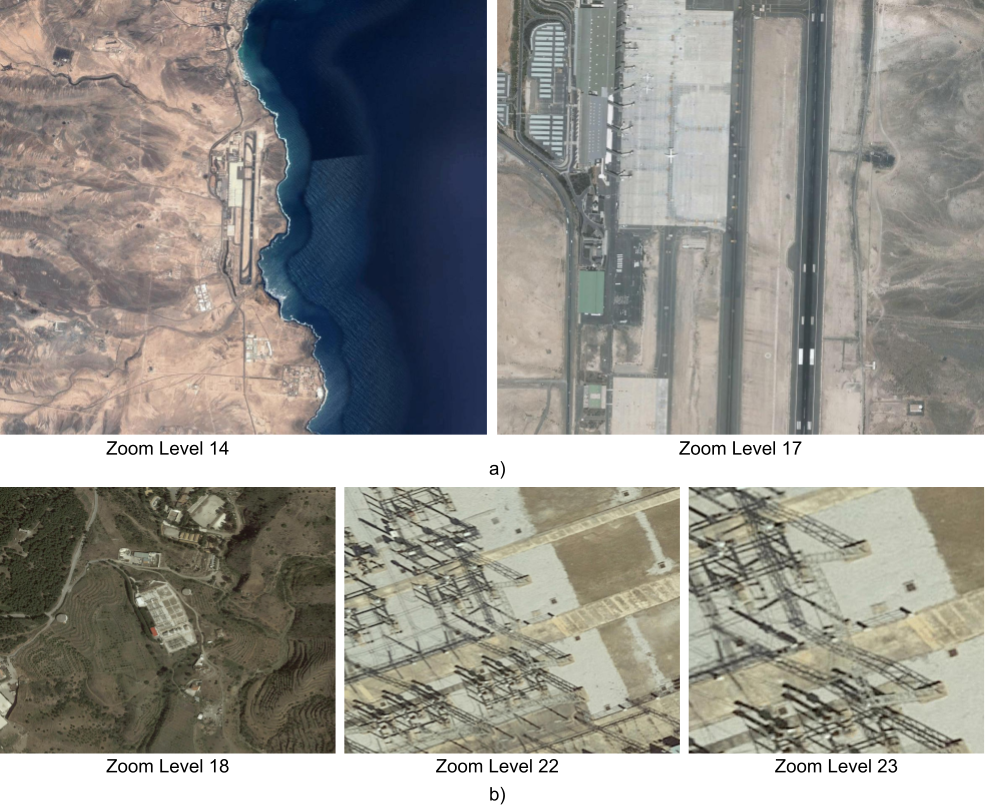}
	\caption{Zoom levels discarded for the test. a) Large scale discard 14 for having the objects too far away and 17 for occupying more of the image. b) Small scale discard 18 for having the objects too far away and 22 and 23 for occupying more of the image.} \label{fig_nozoomlevels}
\end{figure}

Although all selected zoom levels provide useful information for training the detection model, the lowest, 14 and 18, and highest zoom levels, 17 and 22 and 23, require specific manual pre-processing to fit $2000\times2000$ pixels \footnote{Pre-processing includes fusing multiple tiles, cropping a tile and/or resizing the obtained image to $2000\times2000$ pixels. Notice that this size corresponds to the the input layer of the detection model.} so that they can be used for training the detection model. For the test process, no pre-processing is applied and zoom levels 14 and 17 for large scale (Figure \ref{fig_nozoomlevels} (a)) and 18, 22 and 23 for small scale (Figure \ref{fig_nozoomlevels} (b)) infrastructures are discarded. That is, we consider zoom levels in [19,21] for the electrical substation and in [15,16] for the airport class, in the test set. Once the zoom levels are selected for the training process, the images of the target class are downloaded to build subsets CI-SS and CI-LS.

Finally, once the target class dataset is constructed, we analysed all the combinations of zoom levels to determine which one improves the learning process of the detection models. Guided by the performance of the Faster R-CNN on the target class, we discarded the zoom levels that did not help in the learning process of the detector.

\textbf{Small Scale}: 
The initial CI-SS dataset, CI-SS\_train\_alpha, is built using the electrical substation images with zoom levels from 18 to 23. We downloaded 550 images with different zoom levels, as shown in Table \ref{tab_ss_train_alpha}. For building the test set, CI-SS\_test\_alpha, we downloaded 75 images of the electrical substation class with zoom levels from 19 to 21, as shown in Table \ref{tab_ss_test_alpha}. 

\begin{table}[H]
\centering
\footnotesize
	\caption{Number of instances in the electrical substation class, a) CI-SS\_train\_alpha, b) CI-SS\_test\_alpha.}
	\begin{subtable}{.5\linewidth}
	\centering
	\caption{}\label{tab_ss_train_alpha}
	\begin{tabular}{lr}
	\toprule
	Zoom level & Electrical substation \\ 
	\midrule
	18 & 103 \\ 
	19 & 103 \\
	20 & 103 \\ 
	21 & 103 \\
	22 & 103 \\ 
	23 & 103 \\ \hline
	Total & 618 \\
	\bottomrule
	\end{tabular}
	\end{subtable}
	\begin{subtable}{.5\linewidth}
	\centering
	\caption{}\label{tab_ss_test_alpha}
	\begin{tabular}{lr}
	\toprule
	Zoom level & Electrical substation \\ 
	\midrule
	19 & 27 \\
	20 & 27 \\
	21 & 27 \\ \hline
	Total & 81 \\
	\bottomrule
	\end{tabular}
	\end{subtable} 
\end{table}

\textbf{Large Scale}: 
The initial version of CI-LS dataset, CI-LS\_train\_alpha, is built using only airport images with zoom levels from 14 to 17. We downloaded 160 images of airports from Spain and 80 airports from France, as shown in Table \ref{tab_ls_train_alpha}. To build the initial test set, CI-LS\_test\_alpha, we downloaded 32 images of Spanish airports with two zoom levels 15 and 16, as shown in Table \ref{tab_ls_test_alpha}.

\begin{table}[H]
	\centering
	\footnotesize
	\caption{Number of instances for the airport class, a) CI-LS\_train\_alpha, b) CI-LS\_test\_alpha.}
	\begin{subtable}{.5\linewidth}
	\centering
	\caption{}\label{tab_ls_train_alpha}
	\begin{tabular}{lrrr}
	\toprule
	Zoom level & Airport \\ 
	\midrule
	14 & 60 \\ 
	15 & 69 \\
	16 & 251 \\ 
	17 & 124 \\ \hline
	Total & 504 \\
	\bottomrule
	\end{tabular}
	\end{subtable}
	\begin{subtable}{.5\linewidth}
	\centering
	\caption{}\label{tab_ls_test_alpha}
	\begin{tabular}{lr}
	\toprule
	Zoom level & Airport \\ 
	\midrule
	15 & 17 \\
	16 & 16 \\ \hline
	Total & 33 \\
	\bottomrule
	\end{tabular}
	\end{subtable} 
\end{table}

\subsection{Step 2: Extending the dataset with more object classes}

After a careful analysis of the FP committed by the detection model when trained on the initial dataset, we determined all potential object classes that make the detector confuse the target class with other different objects. At this stage, we analysed the impact of each one of these potential FP on the learning of the detector and extended the dataset with more object classes from public datasets. If the performance improves, that potential FP class is maintained in the dataset, otherwise it is eliminated from the dataset. 

For small scale infrastructure, the DOTA dataset will be added since their objects are of similar scales. For large scale infrastructures the DIOR dataset will be used as it contains infrastructures of similar sizes.

\textbf{Small Scale}: Initially, we included in CI-SS\_train\_beta all DOTA classes listed in Table \ref{tab_ss_train_beta}. Then we eliminated each DOTA class one by one and evaluated their impact on the detector performance.

\begin{table}[H]
	\footnotesize
	\centering
	\caption{Number of instances for small scale critical infrastructures, CI-SS\_train\_beta.}
	\label{tab_ss_train_beta}
	\resizebox{\textwidth}{!}{
	\begin{tabular}{l|rrrrrr|r|r}
	\toprule
	Zoom level & 18 & 19 & 20 & 21 & 22 & 23 & DOTA & Total \\
	\midrule
	Electrical substation & 103 & 103 & 103 & 103 & 103 & 103 & - & 618 \\
	Large vehicle & 0 & 3 & 26 & 5 & 3 & 0 & 16923 & 16960 \\
	Swimming pool & 111 & 104 & 62 & 11 & 2 & 0 & 1732 & 2022 \\
	Helicopter & 0 & 0 & 0 & 0 & 0 & 0 & 630 & 630 \\
	Bridge & 19 & 18 & 5 & 0 & 0 & 0 & 2041 & 2083 \\
	Plane & 0 & 0 & 0 & 0 & 0 & 0 & 7944 & 7944 \\
	Ship & 0 & 0 & 0 & 0 & 0 & 0 & 28033 & 28033 \\
	Soccer ball field & 4 & 4 & 1 & 0 & 0 & 0 & 311 & 320 \\
	Basketball court & 0 & 0 & 0 & 0 & 0 & 0 & 509 & 509 \\
	Ground track field & 0 & 0 & 0 & 0 & 0 & 0 & 307 & 307 \\
	Small vehicle & 0 & 0 & 141 & 234 & 68 & 5 & 26099 & 26547 \\
	Harbour & 0 & 0 & 0 & 0 & 1 & 0 & 5937 & 5938 \\
	Baseball diamond & 0 & 0 & 0 & 0 & 0 & 0 & 412 & 412 \\
	Tennis court & 6 & 6 & 1 & 0 & 0 & 0 & 2325 & 2338 \\
	Roundabout & 25 & 26 & 13 & 1 & 0 & 0 & 385 & 450 \\
	Storage tank & 23 & 39 & 36 & 12 & 0 & 0 & 5024 & 5134 \\
	\bottomrule
	\end{tabular}
	}
\end{table}

In addition, as we found that the most relevant new classes are bridge, harbour, storage tank, plane and helicopter, the detector is trained to discriminate these classes too. For building CI-SS\_test\_stable, we included 132 images of the five new classes, as summarised in Table \ref{tab_ss_test_stable}.

\begin{table}[H]
	\centering
	\footnotesize
	\caption{Number of instances in the final version of small scale critical infrastructures, CI-SS\_test\_stable dataset.}
	\label{tab_ss_test_stable}
	\begin{tabular}{lrrrrrr}
	\toprule
		Zoom level & \begin{tabular}[c]{@{}l@{}}Electrical\\ substation\end{tabular} & Helicopter & Bridge & Plane & Harbour & \begin{tabular}[c]{@{}l@{}}Storage\\ tank\end{tabular} \\ 
		\midrule
		19 & 27 & 8 & 21 & 68 & 57 & 136 \\
		20 & 27 & 8 & 15 & 35 & 27 & 50 \\
		21 & 27 & 6 & 13 & 17 & 12 & 24 \\ \hline
		Total & 81 & 22 & 49 & 120 & 96 & 210\\
		\bottomrule
	\end{tabular}
\end{table}

\textbf{Large Scale}: After analysing the FP with Faster R-CNN, we included three object classes from DIOR dataset into CI-LS\_train\_beta, namely train station, bridge and harbour, and built the motorway and industrial area class, see Table \ref{tab_ls_train_beta}. We built a test set, CI-LS\_test\_stable, by including 114 new images of the five classes as it can be seen in Table \ref{tab_ls_test_stable}. 

\begin{table}[H]
	\centering
	\footnotesize
	\caption{Number of instances for large scale critical infrastructures, CI-LS\_train\_beta dataset.}
	\label{tab_ls_train_beta}
	\resizebox{\textwidth}{!}{
	\begin{tabular}{lrrrrrr}
	\toprule
	Zoom level & Airport & Train station & Motorway & Bridge & Industrial & Harbour \\ 
	\midrule
	14 & 60 & 1 & 566 & 1 & 11 & 1 \\
	15 & 69 & 2 & 819 & 1 & 14 & 1 \\
	16 & 251 & 2 & 3207 & 8 & 34 & 1 \\
	17 & 124 & 19 & 2859 & 4 & 50 & 1 \\ \hline
	DIOR & 1327 & 1011 & - & 3967 & - & 5509 \\ \hline
	Total & 1831 & 1035 & 7451 & 3981 & 109 & 5513 \\ \bottomrule
	\end{tabular}
	}
\end{table}

\begin{table}[H]
	\centering
	\footnotesize
	\caption{Number of instances the final version of large scale critical infrastructures, CI-LS\_test\_stable dataset.}
	\label{tab_ls_test_stable}
	\begin{tabular}{lrrrrrr}
	\toprule
		Zoom level & Airport & Train station & Motorway & Bridge & Industrial & Harbour \\ 
		\midrule
		15 & 17 & 25 & 518 & 115 & 59 & 32 \\
		16 & 16 & 22 & 303 & 55 & 27 & 20 \\ \hline
		Total & 33 & 47 & 821 & 170 & 86 & 52 \\
		\bottomrule
	\end{tabular}
\end{table}

\subsection{Step 3: Further increasing the size of the training set}

In this stage, we further increase the number of all the new object classes added to both training subsets using new images from Google Maps.

\textbf{Small Scale}: As the CI-SS\_Det\_beta trained model confuses electrical substation with several elements from urban areas, we included urban areas as context in the new training images in the rest of the classes. Namely, we downloaded a total of 1173 new images. The characteristics of the resulting CI-SS\_train\_stable are shown in Table \ref{tab_ss_train_stable}.

\begin{table}[H]
	\centering
	\footnotesize
	\caption{Number of instances for small scale critical infrastructures, final CI-SS\_train\_stable dataset.}
	\label{tab_ss_train_stable}
	\resizebox{\textwidth}{!}{
	\begin{tabular}{l|rrrrrr|r|rrr|r}
	\toprule
	Zoom level & 18 & 19 & 20 & 21 & 22 & 23 & DOTA & 19 & 20 & 21 & Total \\ \midrule
	Electrical substation & 103 & 103 & 103 & 103 & 103 & 103 & - & 175 & 164 & 144 & 1101 \\
	Swimming pool & 111 & 104 & 62 & 11 & 2 & 0 & 1732 & 807 & 308 & 130 & 3267 \\
	Helicopter & 0 & 0 & 0 & 0 & 0 & 0 & 630 & 20 & 17 & 17 & 684 \\
	Bridge & 19 & 18 & 5 & 0 & 0 & 0 & 2041 & 70 & 34 & 19 & 2206 \\
	Plane & 0 & 0 & 0 & 0 & 0 & 0 & 7944 & 13 & 8 & 2 & 7967 \\
	Soccer ball field & 4 & 4 & 1 & 0 & 0 & 0 & 311 & 142 & 64 & 40 & 566 \\
	Basketball court & 0 & 0 & 0 & 0 & 0 & 0 & 509 & 91 & 49 & 35 & 684 \\
	Ground track field & 0 & 0 & 0 & 0 & 0 & 0 & 307 & 4 & 0 & 0 & 311 \\
	Harbour & 0 & 0 & 0 & 0 & 1 & 0 & 5937 & 1 & 0 & 0 & 5939 \\
	Baseball diamond & 0 & 0 & 0 & 0 & 0 & 0 & 412 & 2 & 0 & 0 & 414 \\
	Tennis court & 6 & 6 & 1 & 0 & 0 & 0 & 2325 & 120 & 45 & 27 & 2530 \\
	Roundabout & 25 & 26 & 13 & 1 & 0 & 0 & 385 & 77 & 25 & 7 & 559 \\
	Storage tank & 23 & 39 & 36 & 12 & 0 & 0 & 5024 & 499 & 213 & 61 & 5907 \\ \bottomrule
	\end{tabular}
	}
\end{table}

\textbf{Large Scale}: We further increased the size of CI-LS\_train\_beta dataset by including 768 new images. The characteristics of the resulting CI-LS\_train\_stable are shown in Table \ref{tab_ls_train_stable}.

\begin{table}[H]
	\centering
	\footnotesize
	\caption{Number of instances for large scale critical infrastructures, final CI-LS\_train\_stable dataset.}
	\label{tab_ls_train_stable}
	\resizebox{\textwidth}{!}{
	\begin{tabular}{lrrrrrr}
	\toprule
	Zoom level & Airport & Train station & Motorway & Bridge & Industrial & Harbour \\ 
	\midrule
	14 & 60 & 5 & 1012 & 37 & 69 & 17 \\
	15 & 69 & 6 & 1280 & 37 & 71 & 17 \\
	16 & 251 & 6 & 3947 & 57 & 116 & 27 \\
	17 & 124 & 27 & 4805 & 168 & 291 & 23 \\ \hline
	DIOR & 1327 & 1011 & - & 3967 & - & 5509 \\ \hline
	Total & 1831 & 1055 & 11044 & 4266 & 547 & 5593 \\ \bottomrule
	\end{tabular}
	}
\end{table}

\section{Experimental study} \label{sec_experimental_study}

This section provides all the performed experimental analysis to obtain CI-dataset and the evaluation of DetDSCI methodology. Section \ref{subsec_considerations} summaries the experimental setup for the analysis. Section \ref{subsec_study_dataset} provides all the detection model results obtained during the CI-dataset construction process. Finally, Section \ref{subsec_study_detdsci} provides the analysis and comparison of the proposed DetDSCI methodology.

\subsection{Experimental setup} \label{subsec_considerations}

The dynamic construction of the dataset requires the use of a good detection model. After a careful experimental analysis, we found that Faster R-CNN is the most suitable for this study as it achieves a good speed accuracy trade-off \cite{huang2017speed}.

For training the detection models, the images were resized to $2000\times2000$ pixels image, which represents the required size of the input layer of modern detectors. A careful selection of the zoom level is necessary so that the entire object can fit in the image.

In the experiments carried out in the next sections, we used Keras \cite{chollet2015keras} as a deep learning framework for classification and TensorFlow \cite{abadi2016tensorflow} as a deep learning framework for detection.

For evaluating and comparing the performance we will use these metrics: \textit{Precision}, \textit{Recall} and \textit{F1}(equation \ref{ec:precrecf1}).

\begin{align}
	\label{ec:precrecf1}
	\begin{split}
	Precision &= \frac{TP}{TP+FP} \\
	Recall &= \frac{TP}{TP+FN} \\
	F1 &= 2 \times \frac{Precision \times Recall}{Precision + Recall}
	\end{split}
\end{align}

where the number of true positives (TP), false positives (FP), and false negatives (FN) is computed for each class.

The detection performance is evaluated in terms of mAP (equation \ref{ec:map}) and mAR (equation \ref{ec:mar}) standard metrics for object detection tasks \cite{lin2014microsoft} given $100$ output regions. 

\begin{align}
	\label{ec:map}
	mAP &= \frac{\sum_{i=1}^{K} AP_{i}}{K} & AP_{i} &= \frac{1}{10} \sum_{r\in[0.5,...,0.95]} \int_{0}^{1}p(r)dr
\end{align}

\begin{align}
	\label{ec:mar}
	mAR &= \frac{\sum_{i=1}^{K} AR_{i}}{K} & AR_{i} &= 2\int_{0.5}^{1}recall(o)do
\end{align}

where given $K$ categories of elements, $p$ represents the precision and r $recall$ defines the area under the interpolated precision-recall curve for each class $i$. Whereas $o$ is IoU (intersection over union) in recall(o) is the corresponding recall under the recall-IoU curve for each class $i$.

The performance of the detection models can be improved with the use of several optimisation techniques, namely data augmentation (DA) and analysing different feature extractors (FE). The eight DA techniques used to this task are listed in Table \ref{table_da_techniques} and their impact will be study on the performance of each detector.

\begin{table}[H]
	\footnotesize
	\centering
	\caption{Data augmentation techniques by model.}
	\label{table_da_techniques}
	\begin{tabular}{ll} \toprule
	Model name & Data augmentation technique \\ \midrule
	DA1 & Normalize image \\
	DA2 & Random image scale \\
	DA3 & Random rgb to gray \\
	DA4 & Random adjust brightness \\
	DA5 & Random adjust contrast \\
	DA6 & Random adjust hue \\
	DA7 & Random adjust saturation \\
	DA8 & Random distort colour \\ \bottomrule
	\end{tabular}
\end{table}

Besides, we consider six feature extractors (FE) listed in Table \ref{table_fe} and train the models with or without the best DA techniques. We will analyse the impact of all these factors on the performance of each detection model.

\begin{table}[H]
\centering
	\footnotesize
	\caption{Configuration of feature extractors for different models.}
	\label{table_fe}
	\begin{tabular}{llll} \toprule
	Model name & Region Proposal & ResNet model & with DA \\ \midrule
	FE1 & Faster R-CNN & ResNet 101 V1 & No \\
	FE2 & Faster R-CNN & ResNet 101 V1 & Yes \\
	FE3 & Faster R-CNN & ResNet 152 V1 & No \\
	FE4 & Faster R-CNN & ResNet 152 V1 & Yes \\
	FE5 & Faster R-CNN & Inception ResNet V2 & No \\
	FE6 & Faster R-CNN & Inception ResNet V2 & Yes \\ \bottomrule
	\end{tabular}
\end{table}

\subsection{Experimental study for the construction of the CI-dataset} \label{subsec_study_dataset}

Section \ref{sec_ci_dataset} provided a detailed description of the construction process of CI-dataset. This subsection provides the experimental results of the detection model at each stage of that process. The performance obtained in steps 1, 2, and 3 are respectively analysed in Section \ref{sssec_analysis_step1}, \ref{sssec_analysis_step2} and \ref{sssec_analysis_step3}. Finally, the experimental analysis of the use of DA techniques and different FE is provided in Section \ref{sssec_improving}.

\subsubsection{Analysis of step 1: Construction of the target class dataset} \label{sssec_analysis_step1}

Once the initial CI-dataset of the target class is constructed, we analysed all the combinations of zoom levels to determine which one improves the learning process of the detection models. Guided by the performance of the detection model on the target class, we discarded the zoom levels that did not help in the learning process of the detector.

\textbf{Small Scale}: The performance of the first detector, CI-SS\_Det\_alpha, trained on different zoom level combinations shows similar results as it can be seen from Table \ref{table_results_substation_zoom_level}. We selected the combination that provides the highest number of images, which is the one that includes all the zoom levels, 18, 19, 20, 21, 22 and 23. 

\begin{table}[H]
	\centering
	\caption{Performance of CI-SS\_Det\_alpha when trained on different zoom level combinations of CI-SS\_train\_alpha and tested on CI-SS\_test\_alpha dataset.}
	\label{table_results_substation_zoom_level}
	\resizebox{\textwidth}{!}{
	\begin{tabular}{lrrrrrr}
	\toprule
	\begin{tabular}[c]{@{}l@{}}Zoom level\\ combination\end{tabular} & Precision & Recall & F1 & \begin{tabular}[c]{@{}r@{}}mAP 0.5\\electrical\\ substation\end{tabular} & \begin{tabular}[c]{@{}r@{}}mAP \\ 0.5-0.95\\ mean\end{tabular} & \begin{tabular}[c]{@{}r@{}}mAR \\ 0.5-0.95\\ mean\end{tabular} \\ 
	\midrule
	\textbf{18,19,20,21,22,23} & \textbf{96,49\%} & 67,90\% & 79,71\% & 87,45\% & 48,30\% & 60,70\% \\
	19,20,21,22,23 & 93,44\% & 70,37\% & \textbf{80,28\%} & 86,23\% & \textbf{51,70\%} & 60,40\% \\
	18,19,20,21,22& 91,94\% & 70,37\% & 79,72\% & \textbf{89,90\%} & 48,70\% & 59,00\% \\
	20,21,22,23 & 92,31\% & 59,26\% & 72,18\% & 79,35\% & 43,50\% & 55,80\% \\
	19,20,21,22 & 89,39\% & \textbf{72,84\%} & 80,27\% & 89,18\% & 51,60\% & \textbf{62,60\%} \\
	21,22,23 & 82,76\% & 29,63\% & 43,64\% & 57,90\% & 28,10\% & 38,40\% \\
	20,21,22 & 89,29\% & 61,73\% & 72,99\% & 80,55\% & 44,50\% & 54,40\% \\
	21,22 & 82,35\% & 17,28\% & 28,57\% & 51,11\% & 24,50\% & 34,70\% \\
	\bottomrule
	\end{tabular}
	}
\end{table}

\textbf{Large Scale}: The performance of the detection model, CI-LS\_Det\_alpha, in different zoom level combinations shows that the best and most stable results are obtained by the combination of these zoom levels, 14, 15, 16 and 17, as it can be seen in Table \ref{table_results_airport_zoom_level}.

\begin{table}[H]
	\centering
	\footnotesize
	\caption{Performance of CI-LS\_Det\_alpha when trained on different zoom level combinations of CI-LS\_train\_alpha and tested on CI-LS\_test\_alpha dataset.}
	\label{table_results_airport_zoom_level}
	\begin{tabular}{lrrrrrr}
	\toprule
		\begin{tabular}[c]{@{}l@{}}Zoom level\\ combination\end{tabular} & Precision & Recall & F1 & \begin{tabular}[c]{@{}l@{}}mAP 0.5\\airport \end{tabular}& \begin{tabular}[c]{@{}l@{}}mAP\\0.5-0.95 \\mean\end{tabular} & \begin{tabular}[c]{@{}l@{}}mAR\\0.5-0.95\\ mean\end{tabular} \\ 
		\midrule
		\textbf{14,15,16,17} & \textbf{87,76\%} & \textbf{86,00\%} & \textbf{86,87\%} & \textbf{89,52\%} & \textbf{61,30\%} & \textbf{69,10\%} \\ 
		14,15,16 & 78,85\% & 82,00\% & 80,39\% & 84,67\% & 55,50\% & 62,10\% \\ 
		15,16,17 & 68,42\% & 78,00\% & 72,90\% & 87,89\% & 54,50\% & 64,20\% \\
		15,16 & 87,23\% & 82,00\% & 84,54\% & 82,66\% & 51,00\% & 57,90\% \\ 
		\bottomrule
	\end{tabular}
\end{table}

\subsubsection{Analysis of step 2: Extending the number of classes} \label{sssec_analysis_step2}

Once the CI-dataset is extended with new classes from public datasets, we analysed whether the new classes improve the performance of the detection models.

\textbf{Small Scale}: As it can be seen from Table \ref{table_dota_erase}, eliminating the three DOTA classes, small vehicle, large vehicle and ship, improves the F1 of CI-SS\_Det\_beta detection model. Therefore, the final dataset CI-SS\_train\_stable contains 13 classes, tennis court, baseball diamond, ground track field, basketball court, soccer-ball field, roundabout and swimming pool in addition to bridge, harbour, storage tank, helicopter, plane and electrical substation. 

\begin{table}[H]
	\centering
	\footnotesize
	\caption{Results of different classes to delete from DOTA dataset trained on CI-SS\_train\_beta and tested on CI-SS\_test\_stable dataset.}
	\label{table_dota_erase}
	\begin{tabular}{lrrr}
	\toprule
		Classes deleted & Precision & Recall & F1 \\ \midrule
		None 	&	88,28	\% & 	58,38	\% & 	70,22	\% \\ 
		- Small vehicle 	&	\textbf{92,61	\%} & 	59,64	\% & 	72,53	\% \\ 
		- Large vehicle 	&	90,30	\% & 	62,44	\% & 	73,81	\% \\ 
		\textbf{- Ship} 	&	90,67	\% & 	\textbf{67,53	\%} & 	77,35	\% \\ 	
		- Tennis court 	&	88,09	\% & 	63,00	\% & 	73,39	\% \\ 	
		- Baseball diamond 	&	89,97	\% & 	66,33	\% & 	76,31	\% \\ 
		- Ground track field 	&	87,02	\% & 	65,77	\% & 	74,84	\% \\ 
		- Basketball court 	&	91,19	\% & 	63,80	\% & 	74,99	\% \\ 
		- Soccer-ball field 	&	93,47	\% & 	66,64	\% & 	\textbf{77,74	\%} \\ 	
		- Roundabout 	&	90,48	\% & 	65,28	\% & 	75,70	\% \\ 
		- Swimming pool 	&	90,74	\% & 	66,55	\% & 	76,73	\% \\ 	
		\bottomrule
	\end{tabular}
\end{table}

\textbf{Large Scale}: The results of the detection model, CI-LS\_Det\_beta, trained on CI-LS\_train\_beta, are shown in Table \ref{table_results_adversarial_airport1}. As it can be observed from this table, including some DIOR classes increases the mAP of the detection model on the airport class to 85,73\%.

\begin{table}[H]
	\centering
	\footnotesize
	\caption{Performance of CI-LS\_Det\_beta when trained on CI-LS\_train\_beta and tested on CI-LS\_test\_stable.}
	\label{table_results_adversarial_airport1}
	\begin{tabular}{llr}
	\toprule
		& & CI-LS\_Det\_beta \\ 
		\midrule
		\multirow{7}{*}{mAP 0.5} & Mean & 22,03\% \\ \cline{2-3} 
		& Airport & \textbf{85,73\%} \\
		& Train station & 6,98\% \\
		& Motorway & 4,30\% \\
		& Bridge & 31,97\% \\
		& Industrial & 2,87\% \\
		& Harbour & 0,31\% \\ \hline
		\multirow{4}{*}{mAP 0.5-0.95} & Mean & 12,20\% \\ \cline{2-3} 
		& Small & 2,00\% \\
		& Medium & 4,70\% \\
		& Large & 14,40\% \\ \hline
		\multicolumn{2}{l}{mAR 0.5-0.95} & 22,10\% \\
		\bottomrule
	\end{tabular}
\end{table}

\subsubsection{Analysis of step 3: Increasing the size of the dataset} \label{sssec_analysis_step3}

Once the final classes are determined, new images are included to further improve the performance of the models.

\textbf{Small Scale}: A comparison between CI-SS\_Det\_beta and the new CI-SS\_Det\_stable, trained on the CI-SS\_train\_stable (Table \ref{tab_ss_train_stable}), tested on the CI-SS\_test\_stable (Table \ref{tab_ss_test_stable}) dataset, is shown in Table \ref{table_results_dota_substation}. The performance of CI-SS\_Det\_alpha trained and tested only on the electrical substation is included in the table as reference as well. These results show clearly that the performance of CI-SS\_Det\_stable improves when increasing the size of the training dataset.

\begin{table}[H]
	\centering
	\footnotesize
	\caption{Performance of CI-SS\_Det\_beta and CI-SS\_Det\_stable, trained on CI-SS\_train\_stable, tested on CI-SS\_test\_stable. CI-SS\_Det\_alpha is trained and tested only on the electrical substation class.}
	\label{table_results_dota_substation}
	\resizebox{\textwidth}{!}{
	\begin{tabular}{llrrr}
	\toprule
		& & \begin{tabular}[c]{@{}l@{}}CI-SS\_Det\_alpha\\ (only ele. sub.)\end{tabular} & \begin{tabular}[c]{@{}l@{}}CI-SS\_Det\_beta\\ (six classes)\end{tabular} & \textbf{\begin{tabular}[c]{@{}l@{}}CI-SS\_Det\_stable\\ (six classes)\end{tabular}} \\ 
		\midrule
		\multirow{7}{*}{mAP 0.5} & Mean & \textbf{87,45\%} & 54,21\% & 65,98\% \\ \cline{2-5} 
		& Electrical substation &\textbf{ 87,45\%} & 78,88\% & 85,00\% \\
		& Plane & 0,00\% & 82,94\% & \textbf{85,30\%} \\
		& Helicopter & 0,00\% & \textbf{33,83\%} & 10,39\% \\
		& Bridge & 0,00\% & 18,33\% & \textbf{63,16\%} \\
		& Storage tank & 0,00\% & 83,07\% & \textbf{92,28\%} \\
		& Harbour & 0,00\% & 58,66\% & \textbf{59,75\%} \\ \hline
		\multirow{4}{*}{mAP 0.5-0.95} & Mean & \textbf{48,30\%} & 32,30\% & 38,60\% \\ \cline{2-5} 
		& Small & 0,00\% & 15,30\% & \textbf{25,90\%} \\
		& Medium & \textbf{31,80\%} & 23,50\% & 27,90\% \\
		& Large & \textbf{49,70\%} & 36,80\% & 43,40\% \\ \hline
		\multicolumn{2}{l}{mAR 0.5-0.95} & \textbf{60,70\%} & 47,80\% & 53,10\% \\ \bottomrule
	\end{tabular} }
\end{table}

For a further analysis, we analysed the TP, FP, FN, Precision, Recall and F1 as shown in Table \ref{table_fp_dota_substation}. As it can be observed, CI-SS\_Det\_stable reduces substantially the number of FP and achieves the best F1 value. Therefore, the CI-SS\_Det\_stable model will be used in the rest of the paper as it provides the highest performance on our target class, electrical substation.

\begin{table}[H]
	\centering
	\footnotesize
	\caption{TP, FP, FN, Recall, Precision and F1 in CI-SS\_test\_stable. CI-SS\_Det\_stable is trained on CI-SS\_train\_stable and CI-SS\_Det\_beta is trained on CI-SS\_train\_beta. For comparison purposes, CI-SS\_Det\_alpha is trained only on airports.}
	\label{table_fp_dota_substation}
	\begin{tabular}{lrrrrrr}
	\toprule
		& TP & FP & FN & Precision & Recall & F1 \\ 
		\midrule
		CI-SS\_Det\_alpha(only ele. sub.) & 117 & 449 & 7 & 20,67\% & \textbf{94,35\%} & 33,91\% \\\hline
		CI-SS\_Det\_beta(six classes) & 75 & 124 & 49 & 37,69\% & 60,48\% & 46,44\% \\
		\textbf{CI-SS\_Det\_stable(six classes)} & 112 & 62 & 12 & \textbf{64,37\%} & 90,32\% & \textbf{75,17\%} \\
		\bottomrule
	\end{tabular}
\end{table}

\textbf{Large Scale}: A comparison between CI-LS\_Det\_beta and the new CI-LS\_Det\_stable, trained on CI-LS\_train\_stable (Table \ref{tab_ls_train_stable}), tested on CI-LS\_test\_stable (Table \ref{tab_ls_test_stable}) dataset, is shown in Table \ref{table_results_adversarial_airport}. The mAP of CI-LS\_Det\_alpha trained and tested only on the airport class is included in the table as reference as well. As it can be seen from these results, CI-LS\_Det\_stable shows very similar mAP on airports than CI-LS\_Det\_beta but much better mAP on the rest of potential FP.

\begin{table}[H]
	\centering
	\footnotesize
	\caption{Performance of CI-LS\_Det\_stable and CI-LS\_Det\_beta tested on CI-LS\_test\_stable and CI-LS\_Det\_alpha trained and tested only on the airport class.}
	\label{table_results_adversarial_airport}
	\resizebox{\textwidth}{!}{
	\begin{tabular}{llrrr}
	\toprule
	& & \begin{tabular}[c]{@{}l@{}}CI-LS\_Det\_alpha\\ (only airports)\end{tabular} & \begin{tabular}[c]{@{}l@{}}CI-LS\_Det\_beta\\ (six classes)\end{tabular} & \textbf{\begin{tabular}[c]{@{}l@{}}CI-LS\_Det\_stable\\ (six classes)\end{tabular}} \\ \midrule
	\multirow{7}{*}{mAP 0.5} & Mean & \textbf{89,52\%} & 22,03\% & 36,48\% \\ \cline{2-5} 
		& Airport &\textbf{ 89,52\%} & 85,73\% & 85,37\% \\
		& Train station & 0,00\% & 6,98\% & \textbf{26,45\%} \\
		& Motorway & 0,00\% & 4,30\% & \textbf{5,16\%} \\
		& Bridge & 0,00\% & 31,97\% & \textbf{40,53\%} \\
		& Industrial & 0,00\% & 2,87\% & \textbf{20,96\%} \\
		& Harbour & 0,00\% & 0,31\% & \textbf{40,40\%} \\ \hline
		\multirow{4}{*}{mAP 0.5-0.95} & Mean & \textbf{61,30\%} & 12,20\% & 18,80\% \\ \cline{2-5} 
		& Small & 0,00\% & 2,00\% & \textbf{2,40\%} \\
		& Medium & 0,00\% & 4,70\% & \textbf{6,50\%} \\
		& Large & \textbf{61,30\%} & 14,40\% & 23,00\% \\ \hline
		\multicolumn{2}{l}{mAR 0.5-0.95} & \textbf{69,10\%} & 22,10\% & 33,90\% \\ \bottomrule
	\end{tabular} }
\end{table}

A comparison with CI-LS\_Det\_stable trained on CI-LS\_train\_stable and tested on CI-LS\_test\_stable is provided in Table \ref{table_metrics_adversarial_airport}. In general, CI-LS\_Det\_stable provides the highest F1. 

\begin{table}[H]
	\footnotesize
	\centering
	\caption{Comparison of TP, FP, FN, TN, Precision, Recall and F1 of CI-LS\_Det\_stable trained on CI-LS\_train\_stable and tested on CI-LS\_test\_stable with CI-LS\_Det\_beta and CI-LS\_Det\_alpha. CI-LS\_Det\_alpha is trained and tested only on the airport class.}
	\label{table_metrics_adversarial_airport}
	\begin{tabular}{lrrrrrr} \toprule
	& TP & FP & FN & Precision & Recall & F1 \\ \midrule
	CI-LS\_Det\_alpha (only airports) & 29 & 19 & 1184 & 60,42\% & 2,39\% & 4,60\% \\ \hline
	CI-LS\_Det\_beta (six classes) & 236 & 35 & 977 & 87,08\% & 19,46\% & 31,81\% \\
	\textbf{CI-LS\_Det\_stable} (six classes) & 334 & 39 & 879 & \textbf{89,54\%} & \textbf{27,54\%} & \textbf{42,12\%} \\ \bottomrule
	\end{tabular}
\end{table}

\subsubsection{Analysis of the improvement of the detection models} \label{sssec_improving}

The selection of the right DA techniques and FE can surely further improve the performance of the detection model. We consider eight DA techniques listed in Table \ref{table_da_techniques} and study their impact on the performance of each detector. Besides we consider six FE listed in Table \ref{table_fe} and train the models with or without the best DA techniques. We analyse the impact of all these factors on the performance of each detection model.

\textbf{Small scale:} Table \ref{table_da_substation} shows the performance of CI-SS\_Det\_stable when applying individually different DA techniques on CI-SS\_train\_stable. As it can be observed from this table, applying DA8, random distort colour, achieves the best results in this model. 

\begin{table}[H]
	\centering
	\caption{Results of the different models with a DA technique in CI-SS\_train\_stable and CI-SS\_test\_stable.}
	\label{table_da_substation}
	\resizebox{\textwidth}{!}{
		\begin{tabular}{llrrrrrrrr}
		\toprule
			& & DA1 & DA2 & DA3 & DA4 & DA5 & DA6 & DA7 & \textbf{DA8} \\ 
			\midrule
			\multirow{7}{*}{mAP 0.5} & Mean & 22,26\% & 67,85\% & 66,84\% & 68,07\% & 66,45\% & 64,83\% & 64,67\% & \textbf{69,07\%} \\ \cline{2-10} 
			& Electrical substation & 0,01\% & \textbf{84,89\%} & 83,65\% & 83,36\% & 82,35\% & 83,23\% & 82,81\% & 82,30\% \\
			& Plane & 41,34\% & 83,23\% & \textbf{88,72\%} & 88,08\% & 82,35\% & 88,06\% & 85,69\% & 86,70\% \\
			& Helicopter & 0,02\% & 19,82\% & 16,48\% & 14,39\% & 14,99\% & 12,42\% & 10,32\% & \textbf{24,52\%} \\
			& Bridge & 15,83\% & 64,90\% & 61,18\% & \textbf{65,86\%} & 62,84\% & 55,08\% & 60,38\% & 64,96\% \\
			& Storage tank & 64,28\% & 90,25\% & 89,44\% & \textbf{91,66\%} & 91,16\% & 91,29\% & 91,47\% & 89,88\% \\
			& Harbour & 12,11\% & 64,02\% & 61,55\% & 65,05\% & 65,03\% & 58,79\% & 57,32\% & \textbf{66,07\%} \\ \hline
			\multirow{4}{*}{mAP 0.5-0.95} & Mean & 12,80\% & 38,70\% & 39,20\% & 39,30\% & 39,20\% & 38,80\% & 38,40\% & \textbf{39,50\%} \\ \cline{2-10} 
			& Small & 0,00\% & 23,30\% & 14,10\% & 24,40\% & 23,80\% & 21,80\% & \textbf{31,00\%} & 13,50\% \\
			& Medium & 2,60\% & 26,50\% & 25,60\% & 27,50\% & \textbf{28,70\%} & 28,20\% & 26,20\% & 26,60\% \\
			& Large & 18,90\% & 43,70\% & 44,90\% & 44,70\% & 44,30\% & 43,60\% & 43,70\% & \textbf{45,60\%} \\ \hline
			\multicolumn{2}{l}{mAR 0.5-0.95} & 23,50\% & 54,20\% & 54,40\% & 53,50\% & \textbf{54,70\%} & 54,10\% & 52,80\% & 54,20\% \\
			\bottomrule
		\end{tabular}
	}
\end{table}

Table \ref{table_ma_da_substation} shows the impact of the different FE and DA on the performance of CI-SS\_Det\_stable. As it can be seen, in mean, the best mAP is obtained when using FE2. This detection model will be the new CI-SS\_Det\_stable.

\begin{table}[H]
	\centering
	\caption{Results of different FE with or without DA techniques in CI-SS\_train\_stable and CI-SS\_test\_stable.}
	\label{table_ma_da_substation}
	\resizebox{\textwidth}{!}{
	\begin{tabular}{llrrrrrr}
	\toprule
		& & FE1 & \textbf{FE2} & FE3 & FE4 & FE5 & FE6 \\ 
		\midrule
		\multirow{7}{*}{mAP 0.5} & Mean & 65,98\% & \textbf{68,97\%} & 63,16\% & 65,39\% & 65,83\% & 63,96\% \\ \cline{2-8} 
		& Electrical substation & 85,00\% & 85,19\% & 83,05\% & 87,55\% & 82,73\% & \textbf{87,78\%} \\
		& Plane & 85,30\% & 84,43\% & 85,81\% & 80,91\% & \textbf{86,29\%} & 84,96\% \\
		& Helicopter & 10,39\% & 23,14\% & 6,83\% & 12,48\% & \textbf{48,03\%} & 6,23\% \\
		& Bridge & \textbf{63,16\%} & 62,38\% & 48,45\% & 50,31\% & 60,54\% & 39,71\% \\
		& Storage tank & \textbf{92,28\%} & 88,97\% & 91,01\% & 90,89\% & 90,93\% & 91,82\% \\
		& Harbour & 59,75\% & 69,70\% & 63,82\% & 70,22\% & 69,71\% & \textbf{73,29\%} \\ \hline
		\multirow{4}{*}{mAP 0.5-0.95} & Mean & 38,60\% & \textbf{40,20\%} & 36,70\% & 37,60\% & 36,50\% & 37,60\% \\ \cline{2-8} 
		& Small & \textbf{25,90\%} & 13,30\% & 4,70\% & 3,10\% & 2,70\% & 3,90\% \\
		& Medium & 27,90\% & \textbf{29,90\%} & 23,60\% & 21,50\% & 29,70\% & 28,60\% \\
		& Large & 43,40\% & \textbf{46,30\%} & 42,20\% & 44,50\% & 40,70\% & 42,10\% \\ \hline
		\multicolumn{2}{l}{mAR 0.5-0.95} & 53,10\% & \textbf{54,10\%} & 51,20\% & 53,10\% & 50,70\% & 51,30\% \\
		\bottomrule
	\end{tabular}
	}
\end{table}

\textbf{Large Scale:} Table \ref{table_da_airport} shows the performance of CI-LS\_Det\_stable when applying different DA techniques on CI-LS\_train\_stable. These results show that applying DA3, random rgb to gray, achieves the best detection results.

\begin{table}[H]
	\centering
	\footnotesize
	\caption{Results of the different models with a DA technique in CI-LS\_train\_stable and CI-LS\_test\_stable.}
	\label{table_da_airport}
	\resizebox{\textwidth}{!}{
		\begin{tabular}{llrrrrrrrr}
	\toprule
			& & DA1 & DA2 & \textbf{DA3} & DA4 & DA5 & DA6 & DA7 & DA8 \\ 
		\midrule
			\multirow{7}{*}{mAP 0.5} & Mean & 3,61\% & 35,91\% & \textbf{37,11\%} & 36,98\% & 36,62\% & 35,04\% & 36,34\% & 36,98\% \\ \cline{2-10} 
			& Airport & 19,54\% & 85,71\% & 90,31\% & 85,75\% & 90,87\% & \textbf{91,50\%} & 88,18\% & 85,84\% \\
			& Train station & 0,07\% & 20,72\% & \textbf{27,98\%} & 26,12\% & 23,53\% & 15,84\% & 19,50\% & 23,39\% \\
			& Motorway & 0,36\% & 4,89\% & 6,19\% & 5,92\% & 6,36\% & 5,20\% & 5,81\% & \textbf{6,63\%} \\
			& Bridge & 0,35\% & 39,44\% & 37,78\% & 40,44\% & 36,33\% & 35,92\% & 36,35\% & \textbf{45,05\%} \\
			& Industrial & 0,11\% & 17,05\% & 21,02\% & 21,05\% & 15,85\% & 15,53\% & \textbf{22,06\%} & 15,04\% \\
			& Harbour & 1,22\% & \textbf{47,64\%} & 39,37\% & 42,62\% & 46,76\% & 46,24\% & 46,13\% & 45,94\% \\ \hline
			\multirow{4}{*}{mAP 0.5-0.95} & Mean & 1,60\% & 18,50\% & \textbf{19,30\%} & 18,20\% & 18,30\% & 18,50\% & 17,90\% & 17,70\% \\ \cline{2-10}
			& Small & 0,10\% & 3,40\% & 3,00\% & \textbf{7,00\%} & 2,20\% & 3,50\% & 2,30\% & 5,20\% \\
			& Medium & 0,00\% & 6,20\% & \textbf{7,30\%} & 6,60\% & 6,30\% & 6,70\% & 6,30\% & 6,00\% \\
			& Large & 3,00\% & 20,70\% & 22,40\% & 21,10\% & 21,70\% & 20,80\% & 21,50\% & \textbf{23,00\%} \\ \hline
			\multicolumn{2}{l}{mAR 0.5-0.95} & 13,10\% & 34,80\% & 34,50\% & \textbf{35,40\%} & 33,40\% & 34,20\% & 34,50\% & 34,70\% \\
			\bottomrule
		\end{tabular}
	}
\end{table}

Table \ref{table_ma_da_airport} shows the impact of the different FE and DA on CI-LS\_Det\_stable. One can see that FE5 obtains the best performance with Inception ResNet V2 without DA techniques. This model will be the new CI-LS\_Det\_stable in the rest of the paper.

\begin{table}[H]
	\centering
	\footnotesize
	\caption{Results of different FE with or without DA techniques in CI-LS\_train\_stable and CI-LS\_test\_stable.}
	\label{table_ma_da_airport}
	\resizebox{\textwidth}{!}{
		\begin{tabular}{llrrrrrr}
	\toprule
			& & FE1 & FE2 & FE3 & FE4 & \textbf{FE5} & FE6 \\ 
		\midrule
			\multirow{7}{*}{mAP 0.5} & Mean & 36,48\% & 37,52\% & 37,67\% & 38,05\% & \textbf{42,34\%} & 40,98\% \\ \cline{2-8} 
			& Airport & 85,37\% & 86,46\% & 84,03\% & \textbf{87,70\%} & 86,01\% & 87,21\% \\
			& Train station & 26,45\% & 24,17\% & \textbf{34,20\%} & 22,31\% & 27,76\% & 22,43\% \\
			& Motorway & 5,16\% & 5,53\% & 4,80\% & 5,77\% & 5,95\% & \textbf{8,01\%} \\
			& Bridge & 40,53\% & 47,81\% & 36,69\% & 48,86\% & \textbf{57,27\%} & 54,25\% \\
			& Industrial & 20,96\% & 17,43\% & 23,53\% & 17,54\% & \textbf{23,64\%} & 22,38\% \\
			& Harbour & 40,40\% & 43,71\% & 42,78\% & 46,13\% & \textbf{53,41\%} & 51,63\% \\ \hline
			\multirow{4}{*}{mAP 0.5-0.95} & Mean & 18,80\% & 18,30\% & 18,80\% & 18,50\% & \textbf{20,30\%} & 20,10\% \\ \cline{2-8}
			& Small & 2,40\% & 5,70\% & 3,20\% & 6,50\% & \textbf{9,70\%} & 7,70\% \\
			& Medium & 6,50\% & 7,30\% & 6,30\% & 6,70\% & \textbf{8,50\%} & 7,20\% \\
			& Large & \textbf{23,00\%} & 21,60\% & 22,00\% & 22,90\% & 22,50\% & 22,40\% \\ \hline
			\multicolumn{2}{l}{mAR 0.5-0.95} & 33,90\% & 36,30\% & 35,10\% & 35,20\% & 35,20\% & \textbf{37,70\%} \\
			\bottomrule
		\end{tabular}
	}
\end{table}

\subsection{Experimental study of DetDSCI methodology} \label{subsec_study_detdsci}

Once CI-dataset is constructed and the final models are trained on the small and the large scale critical infrastructures, we develop the zoom level classifier for the DetDSCI methodology. The construction of the zoom level classifier is presented in Section \ref{subsec_zoom_level_classifier} and the analysis of DetDSCI methodology is shown in Section \ref{subsec_performance_methodology}.

\subsubsection{Construction of the zoom level classifier} \label{subsec_zoom_level_classifier}

In the first stage of DetDSCI methodology, a zoom level classifier analyses the input image and determines the scale of this input. This stage can be addressed either by identifying the specific zoom level of each input image or by identifying intervals of zoom levels.

In particular, we developed and analysed two classification models, the first one is trained on ten zoom level classes, from 14 to 23, and the second classification model is trained on two zoom level intervals, interval [14,17] and [18,23]. 
Table \ref{table_dataset_classifier} shows the number of images used to train and test these two classification models. The used images were selected from datasets CI-SS\_train\_stable, CI-SS\_test\_stable, CI-LS\_train\_stable and CI-LS\_test\_stable.

\begin{table}[H]
	\centering
	\footnotesize
	\caption{Number of images by zoom level used for training and evaluating the classifiers.}
	\label{table_dataset_classifier}
	\begin{tabular}{l|rrrr|rrrrrr}
		\toprule
		 & 14 & 15 & 16 & 17 & 18 & 19 & 20 & 21 & 22 & 23 \\ \midrule
		Train & 252 & 400 & 1256 & 2984 & 200 & 591 & 1080 & 2268 & 6406 & 663 \\ 
		Test & 19 & 52 & 52 & 19 & 44 & 304 & 304 & 304 & 19 & 19 \\ \bottomrule
	\end{tabular}
\end{table}

The confusion matrix for the classification by individual zoom level is shown in Table \ref{table_cm_classifier_indiv}. The overall accuracy of this model is 68,31\%, which is very low.

\begin{table}[H]
	\centering
	\footnotesize
	\caption{Confusion matrix for the classifier by zoom level individually.}
	\label{table_cm_classifier_indiv}
	\begin{tabular}{l|rrrr|rrrrrr}
	\toprule
	Zoom level & 14 & 15 & 16 & 17 & 18 & 19 & 20 & 21 & 22 & 23 \\ \midrule
	14 & 0 & 13 & 5 & 0 & 0 & 0 & 0 & 0 & 1 & 0 \\
	15 & 0 & 14 & 34 & 2 & 0 & 0 & 0 & 2 & 0 & 0 \\
	16 & 0 & 0 & 25 & 26 & 0 & 0 & 1 & 0 & 0 & 0 \\
	17 & 0 & 0 & 1 & 18 & 0 & 0 & 0 & 0 & 0 & 0 \\ \hline
	18 & 0 & 0 & 0 & 33 & 0 & 8 & 2 & 0 & 1 & 0 \\
	19 & 1 & 0 & 0 & 9 & 0 & 209 & 69 & 12 & 4 & 0 \\
	20 & 0 & 0 & 0 & 0 & 0 & 12 & 224 & 57 & 11 & 0 \\
	21 & 0 & 0 & 0 & 2 & 0 & 1 & 6 & 268 & 25 & 2 \\
	22 & 0 & 0 & 0 & 0 & 0 & 0 & 0 & 2 & 17 & 0 \\
	23 & 0 & 0 & 0 & 0 & 0 & 0 & 0 & 1 & 18 & 0 \\ 
	\bottomrule
	\end{tabular}
\end{table}

The confusion matrix for the classification by interval is shown in Table \ref{table_cm_classifier_group}. This model obtains an accuracy of 96,83\%, which is substantially higher than the classification by individual zoom level. Therefore, we selected this classifier to be included in our DetDSCI methodology.

\begin{table}[H]
	\centering
	\footnotesize
	\caption{Confusion matrix for the classifier by zoom level by group.}
	\label{table_cm_classifier_group}
	\begin{tabular}{l|rr}
	\toprule
	Zoom level & {[}14,17{]} & {[}18,23{]} \\ \midrule
	{[}14,17{]} & 134 & 8 \\
	{[}18,23{]} & 28 & 966 \\ \bottomrule
	\end{tabular}
\end{table}

\subsubsection{Analysis of DetDSCI methodology} \label{subsec_performance_methodology}

In this section, we analyse and compare the performance of DetDSCI methodology against the baseline detectors CI-LS\_Det\_stable and CI-SS\_Det\_stable and a baseline detector, Base\_Det, trained on all the data and zoom levels.

The characteristic of each model is:

\begin{itemize}
	\item \textbf{Base\_Det:} is a Faster R-CNN ResNet 101 V1 trained on all the data at all zoom levels from CI-SS\_train\_stable and CI-LS\_train\_stable.
	\item \textbf{CI-LS\_Det\_stable:} is a Faster R-CNN Inception ResNet V2 trained on the CI-LS\_train\_stable dataset.
	\item \textbf{CI-SS\_Det\_stable:} is a Faster R-CNN ResNet 101 V1 with DA techniques trained on the CI-SS\_train\_stable dataset.
	\item \textbf{DetDSCI methodology:} is the methodology by which each input image is classified by the zoom level classifier and based on the output of this classifier, the detector to be used is selected between CI-LS\_Det\_stable or CI-SS\_Det\_stable. 
\end{itemize}

We tested the four models on the images of the target classes, electrical substation from CI-SS\_test\_stable and airport from CI-LS\_test\_stable. The results in terms of TP, FP, FN, Precision, Recall and F1 are shown in Table \ref{performance_DetDSCI}. 

\begin{table}[H]
	\centering
	\footnotesize
	\caption{Performance comparison between DetDSCI methodology, Base\_Det, CI-LS\_Det\_stable and CI-SS\_Det\_stable when tested on the fusion of CI-SS\_test\_stable and CI-LS\_test\_stable.}
	\label{performance_DetDSCI}
	\begin{tabular}{l|llllll} \toprule
	& TP & FP & FN & Precision & Recall & F1 \\ \midrule
	Base\_Det & 70 & 35 & 44 & 66,67\% & 61,40\% & 63,93\% \\
	CI-LS\_Det\_stable & 27 & \textbf{3} & 88 & \textbf{90,00\%} & 23,48\% & 37,24\% \\
	CI-SS\_Det\_stable & 71 & 32 & 44 & 68,93\% & 61,74\% & 65,14\% \\
	\textbf{DetDSCI methodology} & \textbf{83} & 24 & \textbf{32} & 77,57\% & \textbf{72,17\%} & \textbf{74,77\%} \\ \bottomrule
	\end{tabular}
\end{table}

As it can clearly see from this table, DetDSCI methodology overcomes Base\_Det, CI-SS\_Det\_stable and CI-LS\_Det\_stable in all the aspects by achieving the highest performance. In particular, DetDSCI methodology achieves an improvement in F1 of up to 37,53\%.

\section{Conclusions and future work} \label{sec_conclusions}

The detection of critical infrastructures in satellite images is a very challenging task due to the large scale and shapes differences, some infrastructures are too small, e.g., electrical substations, while others are too large, i.e., airports. This work addressed this problem by building the high quality dataset, CI-dataset, organised into two subsets, CI-SS and CI-LS and using DetDSCI methodology. The construction process of CI-SS and CI-LS was guided by the performance of the detectors on electrical substations and airports respectively. 

DetDSCI methodology is a two-stage based approach that first identifies the zoom level of the input image using a classifier and then analyses that image with the corresponding detection model, CI-LS\_Det\_stable or CI-SS\_Det\_stable. DetDSCI methodology achieves the highest performance with respect to the baseline detectors not only in the target objects but also in the rest of infrastructure classes included in the dataset.

As conclusions, the proposed datasets and methodology are the best solution for addressing the problem of different and dissimilar scale critical infrastructures detection in remote sensing images. This approach can be easily extended to more critical infrastructures. 

As a future work, we will extend the dataset and methodology to more critical infrastructures and design a strategy to group sets of classes according to their zoom level and shared features, with the objective to achieve more robust detection models.

\section*{Acknowledgements}

This work was partially supported by projects P18-FR-4961 (BigDDL-CET) and A-TIC-458-UGR18 (DeepL-ISCO). S. Tabik was supported by the Ramon y Cajal Programme (RYC-2015-18136).

\bibliographystyle{plain}
\bibliography{main}
\end{document}